\documentclass[conference]{IEEEtran}
\IEEEoverridecommandlockouts
\usepackage{cite}
\usepackage{amsmath,amssymb,amsfonts}
\usepackage{algorithmic}
\usepackage{graphicx}
\usepackage{textcomp}
\usepackage{xcolor}
\def\BibTeX{{\rm B\kern-.05em{\sc i\kern-.025em b}\kern-.08em
    T\kern-.1667em\lower.7ex\hbox{E}\kern-.125emX}}
\title{An Ensemble Learning Approach towards Waste Segmentation in Cluttered Environment}

\author{\IEEEauthorblockN{Maimoona Jafar}
\IEEEauthorblockA{\textit{SEECS, NUST}\\
Islamabad, Pakistan \\
mjafar.mscs21seecs@seecs.edu.pk }\\
\IEEEauthorblockN{Ahsan Saadat}
\IEEEauthorblockA{\textit{SEECS, NUST} \\
Islamabad, Pakistan \\
ahsan.sadaat@seecs.edu.pk}
\and
\IEEEauthorblockN{Syed Imran Ali}
\hspace{5cm}
\IEEEauthorblockA{\textit{SEECS, NUST} \\
Islamabad, Pakistan \\
imran.ali@seecs.edu.pk}\\
\IEEEauthorblockN{Shah Khalid}
\IEEEauthorblockA{\textit{SEECS, NUST} \\
Islamabad, Pakistan \\
shah.khalid@seecs.edu.pk}
\and
\IEEEauthorblockN{Muhammad Bilal}
\IEEEauthorblockA{\textit{SEECS, NUST} \\
Islamabad, Pakistan \\
bilal.ali@seecs.edu.pk}
}
\usepackage{url}
\begin{document}

\maketitle

\begin{abstract}

Environmental pollution is a critical global
issue, with recycling emerging as one of the most viable
solutions. This study focuses on waste segregation, a crucial step in recycling processes to obtain raw material. Recent advancements in computer vision have significantly contributed to waste classification and recognition. In waste segregation, segmentation masks are essential for robots to accurately
localize and pick objects from conveyor belts. The complexity
of real-world waste environments, characterized by deformed
items without specific patterns and overlapping objects further
complicates waste segmentation tasks. This paper proposes
an Ensemble Learning approach to improve segmentation
accuracy by combining high performing segmentation models,
U-Net and FPN, in weighted average method. U-Net excels in
capturing fine details and boundaries in segmentation tasks,
while FPN effectively handles scale variation and context in
complex environments, and their combined masks result in more
precise predictions. The dataset used closely mimics real-life
waste scenarios, and pre-processing techniques were applied
to enhance feature learning for deep learning segmentation
models. The ensemble model, referred to as EL-4, achieved
an IoU value of 0.8306, an improvement over U-Net’s 0.8065,
and reduced Dice loss to 0.09019 from FPN’s 0.1183. This
study could contribute to the efficiency of waste sorting at
Material Recovery Facility, facilitating better raw material acquisition for recycling with minimal human intervention and enhancing the overall throughput.\\
  
\end{abstract}


\section{Introduction}
Waste management has become a significant issue in recent years due to increased population and consumerism. It leads to unprecedented levels of waste production. Increase in population leads to increase in consumption of materials which are expected to rise by 2.3\% per year or more \cite{b1}. Changing lifestyles backed by economic growth have led to a rise in consumerism. Production has increased and the disposal of single-use items, packaging materials, and electronic waste is common in everyday life. This trend severely burdens the existing waste management system, on the one hand. On the other hand, it put more pressure on natural resources for new materials to meet the ever growing demand \cite{b2}. Effective waste management is crucial to mitigate these impacts. Conventional methods, such as landfills and burning, do not address the current scale of waste production. The most effective approach involves reducing, reusing, and recycling waste. These reduces the strain on the natural resources that are continuously harvested for new products. Material Recovery Facilities play a crucial role in recycling by retrieving raw materials for production. Traditionally, these facilities rely on human labor. However, there is a growing trend towards utilizing robots. In this regard,efficient and intelligent waste identification and recycling systems are being developed, leveraging advances in artificial intelligence and machine learning to improve the accuracy of the sorting and streamline recycling processes. \\

In computer vision, some tasks are performed for IWIR  in waste segregation includes classification, detection and segmentation. Many machine and deep models have been developed for waste management\cite{b3}, but the field remains highly diverse and complex. Real-time applications need to account for the true nature of waste, which often appears in cluttered and occluded forms rather than in isolation. This clutter can obscure objects, making it difficult for models to identify and classify waste accurately. Additionally, the variety of deformations and the wide variation of sizes of waste items add to the challenge, complicating recognition and processing. While most studies focus on waste classification, they often overlook the fact that multiple waste items can be present in a single image. Object detection methods can generate bounding boxes and labels for multiple objects, but complete automation of waste management requires more detailed information. This has led to a growing trend in waste management research towards segmentation techniques\cite{b4}, which provide pixel-wise classification of input data, offering a more precise and comprehensive approach to handling diverse and cluttered waste streams.

This paper aims to address the challenges in waste segmentation at MRF. Annotated waste stream pictures acquired from MRF are employed in this study. This is the most useful image type for categorizing waste items,  the ZeroWaste-f dataset \cite{b5} which was created specifically with this aim. The purpose of this study is to create a deep segmentation model that can generate predictive masks for the waste items on conveyor belt at MRF.\\

The goal of this study is to propose a segmentation model that can generate accurate predictive masks for waste items in cluttered environment. The key objectives are listed below:\\

\begin{enumerate}
    \item Determine the high performance segmentation models in cluttered waste based on accuracy.
    \item Formulate a method to combine high performance models to obtain highest segmentation accuracy.
    \item Explore various encoders for extracting features from dataset which minimizes computational cost.\\
\end{enumerate}

Based on the defined objectives, a series of experiments were conducted. As a result, the following key contributions were made:\\

\begin{enumerate}
        \item Evaluated seven state-of-the-art deep segmentation models (U-Net\cite{b6}, U-Net++\cite{b7}, Feature Pyramid Network (FPN)\cite{b8}, Multi-Attention Network (MANet)\cite{b9}, LinkNet\cite{b10}, Pyramid Scene Parsing Network (PSPNet)\cite{b11} and Pyramid Attention Network (PAN)\cite{b12}) on the ZeroWaste-f dataset, demonstrating their performance. Identified U-Net and FPN as high performance models, with high IoU score and low Dice Loss, accurately identifying the waste objects.

        \item Proposed an ensemble model combining U-Net and FPN through equal weighted averaging mechanism to enhance segmentation accuracy.
        
        \item Utilized EfficientNet\cite{b13} encoders (B0, B1, B2, B3, and B4) as backbone network, creating five Ensemble Learning (EL) model variants (EL-0, EL-1, EL-2, EL-3 and EL-4) to reduce computational resource utilization.\\
\end{enumerate}

Section 2 reviews the relevant literature. Section 3 elaborates the methodology adopted to obtain the proposed model. Section 4 gives the evaluations of the model on the dataset. Section 5 is concludes the paper.

\section{Literature Review}

This study focuses on sorting waste at centralized locations, such as MRF. However some other instances where some effective segmentation model for waste is being proposed are also analyzed. Evolution of computer vision in waste sorting, its prospects and challenges have been reviewed. The computer vision algorithms used for waste sorting, conventional and deep learning methods, have been explored with specific focus on segmentation in cluttered environment.\\

CV algorithms have been adapted in the field of waste sorting, which includes steps of image processing, feature extraction and ML algorithms for classification. Koyanaka et al. \cite{b14} develops an automatic sorting method for light weight metal scrap using a 3D imaging camera and processes and classifies through multi-variate analysis. Leitner et al. \cite{b15} proposes a method for real-time classification of waste polymers in a prototype of an automated industrial sorting facility based on Near-infrared (NIR) spectroscopy employing dissimilarity-based classifier (DBC). Tachwali et al. \cite{b16} puts forward real time automatic plastic bottle sorting system based on NIR spectroscopy and quadratic discriminant function fused with decision tree classifiers. Gomes et al. \cite{b17} proposes a methodology for the classification of fine particles from Construction and Demolition Waste based on scanning electron microscopy (SEM) and image analysis. Nawrocky et al. \cite{b18} presents technique to sort PET, type of plastic and poly-coat materials employing intensity images of these objects using image histograms as input to a SVM which performs classification. Özkan et al. \cite{b19} provides automatic classification scheme using SVM for plastic bottles types which classifies  for PET and non-PET plastic types. \\

\begin{figure*}[!ht]
\centerline{
\includegraphics[width=7in]{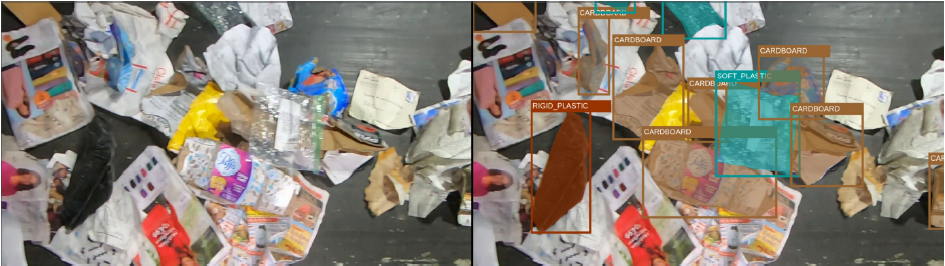}}
\caption{Zerowaste-f dataset segmentation using Mask-RCNN}
\label{fig}
\end{figure*}

ML was widely used for classification purposes but it is deficient of the ability to obtain high level representations from input images. It is slowly being replaced by Deep Learning which has the ability to extract features through series of connected layers. When considering automating the environment of MRF, the task like robotic grasping of an object is of primal importance. In this regard, focus has been shifting towards segmentation. Segmentation classifies each pixel. In this way, it helps to identify the shape of each object and generate pixel wise object mask and thus prove more helpful at MRF. \\

A multi-stage approach for waste segmentation is proposed in \cite{b21} first segmenting the region and then the object on self-proposed dataset MJU using Resnet101 backbone. \cite{b22}  develops a waste segmentation technique based on Encoder-Decoder method of SegNet \cite{b23} architecture using CNN with varying number of filters in each convoultion layer and evaluated model on TrashNet dataset.  works for the problem of road garbage classification and segmentation for measuring the road cleanliness index based on semantic segmentation (CISS) with simple Deep supervision UNet++ (DUNet++) which is UNet with four deep supervision layers added behind the up-sampling layers is proposed. \cite{b24}investigate waste detection and classification in an industrial sorting facility. WaRP (Waste Recycling Plant) dataset \cite{b25} is used for detection and for classification same dataset is preprocessed to obtain images with labels. Various deep learning classification architecture such as CNN, AlexNet, VGG16 and others are tested on WaRP-C. The highest scores were recorded of ConvNeXt \cite{b26} and EfficientNet-B5 models, while the ResNet-18 model took the least time. For waste detection, several object detection models including YOLO were evaluated on WaRP-D and highest precision was recorded by Task-aligned One-Stage Object Detection \cite{b27}. \\

It has been observed that many studies have used dataset that does not truly capture the  state of waste material in MRFs. MRFs receives tons of garbage per day which needs to be sorted for recycling with least amount of contamination or mislabeling efficiently. In this regard, such segmentation models have been studied which were designed specifically for the problem of clutter for segmentation task. Construction Waste presents the challenges of the mixture and clutter nature which includes a wide range of materials (e.g., rock, stone, rubble, debris, concrete, and bricks, gravel, wood and packaging). DeepLabV3+\cite{b28} with Xception \cite{b29} and ResNet backbone separately pretrained on ImageNet dataset used by \cite{b30} to show the efficiency.  \cite{b31} introduces the RGB-DL depth fusion strategy for identification of recyclables from cluttered and assorted Construction and Demolition Waste  streams in MRF which uses late integration of depth features at the end of the segmentation pipeline improving precision. \cite{b32} proposed a one-shot segmentation model, named MaskNet, to identify and segment object seen first time with no prior training data on cluttered Omniglot dataset. The model MaskNet includes the U-Net architecture for segmentation and Siamese Neural Network embedding for detection. It improves segmentation and localization. \cite{b5} investigated recyclable waste detection in severely distorted and transparent objects in cluttered scenes. They introduced the ZeroWaste dataset for industrial-grade waste detection and segmentation, conducting experiments with various CV models. RetinaNet \cite{b33}, Mask R-CNN \cite{b34}, and TridentNet \cite{b35}were evaluated for object detection, with TridentNet giving the highest precision. For segmentation, DeepLabv3+ using a ResNet101 backbone was giving the highest score of 0.5213. The main limitation which has been mentioned is that these architectures does not perform well in cluttered environment. This dataset has been used in present study because of it represents true nature of the waste as in used paper recovery for recycling in a MRF.

\section{Methodology}

For segmentation of ZeroWaste-f dataset, the model designed in study consists of multiple modules, which includes preprocessing steps, feature extraction through encoder, processing using various deep learning models to obtain segmentation masks, combination of segmentation masks to obtain aggregate mask and then using final segmentation mask to obtain pixel level classification. These steps are explained in more detail in the below sections.

\subsection{Dataset}\label{AA}
ZeroWaste-f is publicly available waste segmentation and detection dataset which is largest at the present in this domain. This dataset was particularly gathered for evaluating labelled automated waste detection and segmentation. Specific traits of this dataset are that images are highly cluttered and objects are deformed as they are present in the waste. Certain objects which are hard to distinguish are also differentiated through extensive labelling. 

\begin{figure*}[!ht]
\centerline{
\includegraphics[width=7in]{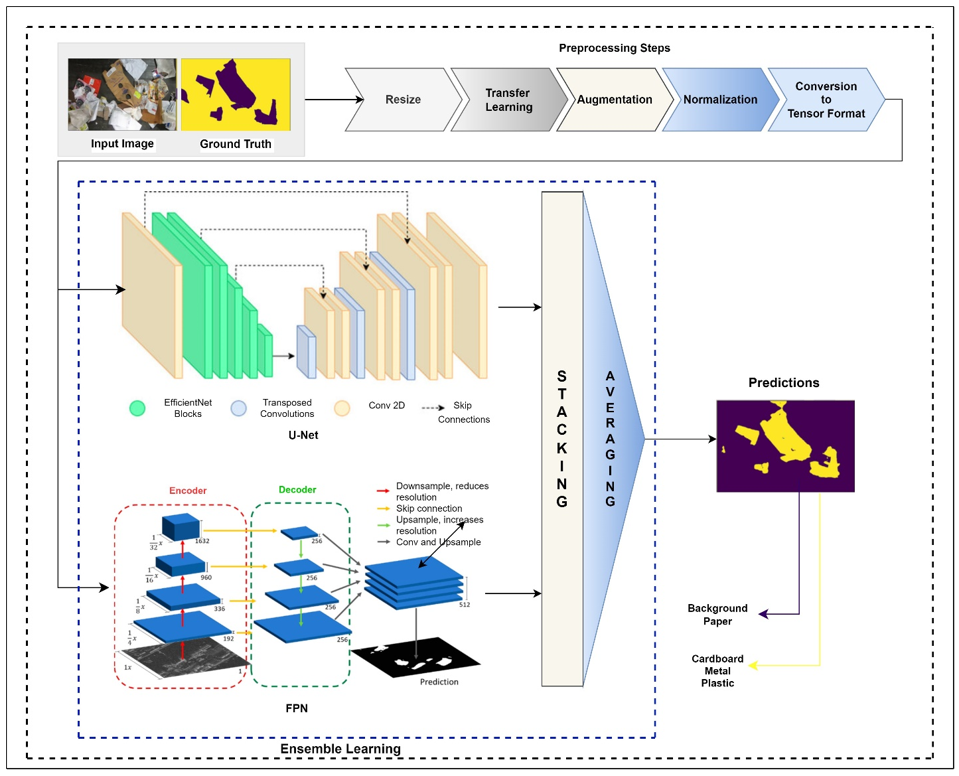}}
\caption{Workflow of the EL Model for Segmentation over ZeroWaste-f dataset}
\label{fig}
\end{figure*}

\subsection{Pre-processing}\label{AA}

In the pre-processing, images are resized since they are present in full HD form in dataset which requires extensive computing resources and time. The new dimensions of image are defined as 320 x 480. Image augmentation is performed on the train dataset with the aim of diversifying it. It includes flipping, scaling, rotating, adding noise, perspective transforms, and color adjustments. Transfer learning is used since it speeds up the learning process as the model has already learned the generic features from the large dataset. Model's encoder uses the pre-trained weights learned over the ImageNet\cite{b36} datatset. Normalization is applied according to ImageNet.First pixel values are scaled for all images then mean and standard deviation are calculated for each color channel separately through the entirety of dataset.

\subsection{Segmentation Models}\label{AA}
State-of-the-art segmentation models were evaluated over the ZeroWaste-f dataset and results were noted. All these experiments were performed with the help of Segmentation Models Pytorch \cite{b37}. With the help of SMP, seven different segmentation models were applied on the ZeroWaste-f dataset namely, U-Net, U-Net++, MANet, LinkNet, FPN, PSPNet and PAN. Their results are discussed in Section 4.

\subsection{Proposed Model}\label{AA}
In this research, ensemble learning provides the basis for the proposed segmentation model. Through ensemble learning, U-Net and FPN are combined in a way to obtain optimal segmentation score over ZeroWaste-f dataset.\\ 

While experimenting with their approach of combination, different schemes were tried including element wise addition of segmentation masks from both models, concatenation and weighted average. After some experiments it was observed that the stacking and averaging approach was giving high performance as compare to all other methods, therefore it was selected as final approach.\\

The workflow of the ensemble model is depicted in Figure 2. Preprocessing is applied to the training data. The preprocessed data is then fed to both models in parallel. Both U-Net and FPN use EfficientNet as the encoder to extract features from the input. EfficientNet is employed in the U-Net encoder during downsampling, replacing ResNeXt's convolutional blocks with its inverted residual blocks, which reduces computational cost while enhancing segmentation performance. In the FPN architecture, EfficientNet changes the feature extraction blocks, efficiently capturing multi-scale features and complementing FPN's performance in segmentation tasks. The segmentation masks generated by U-Net and FPN are combined using a stacking and averaging method. This ensemble learning approach leverages the diversity of predictions from individual models, enhancing the final prediction's overall accuracy and reliability.

\subsection{Training Process}\label{AA}

The ensemble learning model for segmentation was trained with EfficientNet encoder pre-trained on ImageNet dataset. Classes were specified with input dataset and their truth segmentation masks. The output layer of the models, U-Net and FPN uses activation function which based on the task of multiclass segmentation, softmax2d was specified. The model was trained on the dataset using a Tensor T4 GPU on Google Colab with CUDA support. Preprocessing steps are applied on the training data and the training and validation datasets are converted into batches of data suitable for training and validation. For training dataset, batch-size 8 is specified. Accordingly, there will be 376 batches during training. Shuffling of data within batches during training is disabled. Validation dataset has batch-size 1, to evaluate each image individually. Loss function evaluated during training is dice loss which is commonly used for image segmentation tasks, measuring the overlap between predicted and target masks. IoU (Intersection over Union), is used as segmentation metric, with a threshold of 0.5, to measure the similarity between predicted and ground truth masks. It's a standard metric for evaluating segmentation accuracy. Adam optimizer with learning rate of 0.0001 is utilized for convergence during training. 

\section{Results}
EL model based on U-Net and FPN, used for segmentation of waste in cluttered and diverse environment is implemented in Google Colab with help of Pytorch Framework. The model was trained using an NVIDIA Tesla T4 GPU with 16 GB GDDR6 memory and 12 GB RAM on Google Colab with CUDA support. Pytorch framework is used to build and train neural networks through Timm, the Pytorch Image Models library, which provides their implementation. Specific Implementation of deep learning models for semantic segmentation as provided by segmentation models pytorch is utilized. In ZeroWaste-f dataset, images are divided into training, validation and test splits with ratio of 67:13:20. 

\subsection{State-of-the-art Segmentation Models Performance}
In this study, IoU score, dice loss, f1 score and segmentation masks have been used for evaluation of the deep learning segmentation models. These are the bench mark criteria of measuring the performance of segmentation models.  

\begin{table}[htbp]
\caption{IoU Scores of State-of-the-art Segmentation Models}
\begin{center}
\renewcommand{\arraystretch}{1.5} 
\begin{tabular}{|c|c|c|c|}
\hline
\textbf{Model} & \multicolumn{3}{|c|}{\textbf{IoU Score}} \\
\cline{2-4} 
\textbf{} & \textbf{\textit{Train}} & \textbf{\textit{Valid}} & \textbf{\textit{Test}} \\
\hline
UNet & 0.7011 & 0.7724 & 0.8065 \\
\hline
UNet++ & 0.6577 & 0.7480 & 0.7798 \\
\hline
MANet & 0.6373 & 0.6966 & 0.7324 \\
\hline
LinkNet & 0.5560 & 0.5941 & 0.6608 \\
\hline
FPN & 0.7738 & 0.7687 & 0.7953 \\
\hline
PSPNet & 0.7348 & 0.7392 & 0.7715\\
\hline
PAN & 0.7310 & 0.7162 & 0.7473\\
\hline
\end{tabular}
\label{tab1}
\end{center}
\end{table}

The results in Table I shows that U-Net achieves the highest IoU score of 0.8065 followed by FPN achieving 0.7953. These scores depicts model’s effective feature representation. PSPNet and PAN gives competitive IoU scores. PAN performs well but slightly lower compared to FPN and PSPNet, particularly on the validation set. U-Net++ gives score competitive to U-Net especially on validation and test sets. MANet gives moderate IoU scores. LinkNet has the lowest IoU scores among the models, indicating less accurate segmentation performance compared to the others. This could be interpreted that in the segmentation task U-Net is proving itself effective while FPN also shows strong performance. This suggests their suitability for segmentation task over this dataset.\\

\begin{table}[htbp]
\caption{Dice Loss of State-of-the-art Segmentation Models}
\begin{center}
\renewcommand{\arraystretch}{1.5} 
\begin{tabular}{|c|c|c|c|}
\hline
\textbf{Model} & \multicolumn{3}{|c|}{\textbf{Dice Loss}} \\
\cline{2-4} 
\textbf{} & \textbf{\textit{Train}} & \textbf{\textit{Valid}} & \textbf{\textit{Test}} \\
\hline
UNet & 0.3829 & 0.2273 & 0.2084 \\
\hline
UNet++ & 0.3966 & 0.2481 & 0.2308 \\
\hline
MANet & 0.3469 & 0.2334 & 0.2080 \\
\hline
LinkNet & 0.4525 & 0.3830 & 0.2080 \\
\hline
FPN & 0.1316 & 0.1351 & 0.1183 \\
\hline
PSPNet & 0.1634 & 0.1565 & 0.1349\\
\hline
PAN & 0.1835 & 0.1754 & 0.1577\\
\hline
\end{tabular}
\label{tab1}
\end{center}
\end{table}

Table II shows the results of the Dice Loss which calculated the dissimilarity. So its lower values are associated with good performance. According to this, FPN has lowest Dice Loss (0.1183) and hence predicting the masks most similar to the ground truth with least amount if difference. On the whole, FPN, PSPNet and PAN, exhibit lower Dice Loss values, especially on the Test dataset which suggests that the segmentation masks produced by them are closer to ground truth. This also indicates that model which uses pyramid level scheme during feature extraction have lower Dice Loss. , U-Net shows moderate levels of dice loss value. U-Net++, MANet and LinkNet have higher Dice Loss values compared to some other. FPN, PSPNet and PAN demonstrates better performance with lower Dice Loss values, suggesting they effectively capture the segmentation details and generalize well across different datasets.\\

U-Net had the highest IoU score among all the segmentation models evaluated, while FPN achieved the lowest Dice Loss on the same dataset. Based on these results, U-Net and FPN were selected to be combined in an ensemble learning (EL) model. The rationale behind this selection is that these two models complement each other in the best possible way. U-Net's strength in achieving high IoU scores and FPN's ability to maintain low Dice Loss values signifying its superior performance in minimizing the dissimilarity between the predicted and actual segmentation masks. By leveraging the strengths of both models, the ensemble learning approach aims to produce more accurate and reliable segmentation results on the ZeroWaste-f dataset.

\subsection{Ensemble Learning Model Performance}

There are different variants of this EL model using different types of EfficientNet encoders, B0, B1, B2, B3, and B4, and named likewise. The IoU score and Dice Loss computed using these models over this dataset and recorded in Table IV, Table V and Table VI.\\

\begin{table}[htbp]
\caption{IoU Scores of EL based Segmentation Models}
\begin{center}
\renewcommand{\arraystretch}{1.5} 
\begin{tabular}{|c|c|c|c|}
\hline
\textbf{Model} & \multicolumn{3}{|c|}{\textbf{IoU Score}} \\
\cline{2-4} 
\textbf{} & \textbf{\textit{Train}} & \textbf{\textit{Valid}} & \textbf{\textit{Test}} \\
\hline
EL-0 & 0.8531 & 0.7910 & 0.8147 \\
\hline
EL-1 & 0.8594 & 0.8041 & 0.8196 \\
\hline
EL-2 & 0.8681 & 0.8025 & 0.8282 \\
\hline
EL-3 & 0.8778 & 0.8058 & 0.8235 \\
\hline
EL-4 & 0.8802 & 0.8091 & 0.8306 \\
\hline
\end{tabular}
\label{tab1}
\end{center}
\end{table}

The EL model designed by combining segmentation masks from U-Net and FPN while using EfficientNet encoder was trained and tested over ZeroWaste-f dataset. The IoU scores for all EL models on the test set are high, indicating that all ensemble models are performing well in terms of segmentation accuracy. Table 4 shows EL-4 has the highest IoU score on the training set (0.8802), showing it can capture fine details. On the test set, EL-4 achieves the highest IoU on the test set (0.8306) among other variants, representing the best generalization and performance across unseen data. EL-2 and EL-3 also perform very well, with slight differences in IoU scores compared to EL-4. These models could be considered as alternatives if computational efficiency or training time is a concern. EL-4 is the top performer across all datasets (train, validation, and test), indicating that the combination of U-Net and FPN with the EfficientNet-B4 encoder yields the best segmentation results.\\

\begin{table}[htbp]
\caption{Dice Loss of EL based Segmentation Models}
\begin{center}
\renewcommand{\arraystretch}{1.5} 
\begin{tabular}{|c|c|c|c|}
\hline
\textbf{Model} & \multicolumn{3}{|c|}{\textbf{Dice Loss}} \\
\cline{2-4} 
\textbf{} & \textbf{\textit{Train}} & \textbf{\textit{Valid}} & \textbf{\textit{Test}} \\
\hline
EL-0 & 0.08279 & 0.1211 & 0.1068 \\
\hline
EL-1 & 0.07901 & 0.1126 & 0.1034 \\
\hline
EL-2 & 0.07379 & 0.1133 & 0.0977 \\
\hline
EL-3 & 0.06846 & 0.1114 & 0.1012 \\
\hline
EL-4 & 0.06713 & 0.1092 & 0.09019 \\
\hline
\end{tabular}
\label{tab1}
\end{center}
\end{table}

Dice Loss values for all EL models, exhibiting good performance. Dice Loss decreases consistently from EL-0 (0.08297) to EL-4 (0.06713), showing that higher EfficientNet variants improve segmentation accuracy during training. The improvement in validation Dice Loss suggests that higher variants of EfficientNet encoders (particularly EL-4) generalize better to validation data, with EL-4 achieving the lowest validation Dice Loss. Dice Loss on the test set also improves with higher variants: EL-0 (0.1068) to EL-4 (0.09019). EL-4 achieves the lowest Dice Loss on the test set (0.09019), demonstrating the best generalization and performance on unseen data. \\

EL-4 is the top performer across all datasets (train, validation, and test) in terms of both IoU and Dice Loss, indicating that the combination of U-Net and FPN with the EfficientNet-B4 encoder yields the best segmentation results. EL-2 and EL-3 also perform very well, with slight differences in Dice Loss values compared to EL-4. These models could be considered as alternatives if computational efficiency or training time is a concern.\\

The segmentation masks were generated and studied across ground truth to realize that complexity of the domain and indefinite and multitude of shapes is also affecting the prediction of masks. Still it is generating masks very close to ground masks as could be observed in Figure 3.\\

\section{Conclusion}
In this study, the state-of-the-art segmentation models have achieved high score, like U-Net scored 0.8065, which is mainly attributed to the pre-processing applied on the dataset. An EL model based on U-Net and FPN are ensemble which shows significant improvement for waste segmentation in cluttered environment. It increase the IoU value to 0.8306 for EL-04 from 0.8065 for U-Net. Dice loss was reduced from this 0.1183 for FPN to 0.09019 for EL-04. This study is expected to be beneficial towards the problem of sorting waste at MRF. It will help in obtaining better raw materials for recycling products with limited human interference and increase the efficiency and throughput of the MRF. \\

\end{document}